\documentclass[9pt]{article}

\usepackage{geometry}
\usepackage{graphicx} 
\usepackage{subfigure} 
\usepackage{natbib}

\usepackage{algorithm}
\usepackage{algorithmic}

\usepackage{epstopdf}
\usepackage{times,bm}
\usepackage{algorithm}
\usepackage{algorithmic}[1]
\usepackage{graphicx}
\usepackage{subfigure,amssymb,mathrsfs,amsthm,amsfonts,amsmath}
\usepackage{amsmath,graphicx,caption,color,epsfig,float,pbox,tabularx,wrapfig,mathrsfs,multirow,array}

\newcommand{\bfA}{\mathbf{A}}

\newcommand{\bfB}{\mathbf{B}}

\newcommand{\bfD}{\mathbf{D}}
\newcommand{\bfL}{\mathbf{L}}
\newcommand{\bfQ}{\mathbf{Q}}
\newcommand{\bfP}{\mathbf{P}}
\newcommand{\bfu}{\mathbf{u}}
\newcommand{\bfM}{\mathbf{M}}

\newcommand{\bfI}{\mathbf{I}}

\newcommand{\bfS}{\mathbf{S}}

\newcommand{\bfK}{\mathbf{K}}
\newcommand{\bfU}{\mathbf{U}}
\newcommand{\bfV}{\mathbf{V}}

\newcommand{\bfY}{\mathbf{Y}}

\newcommand{\bfv}{\mathbf{v}}
\newcommand{\bfzero}{\mathbf{0}}

\newcommand{\bfSigma}{\mathbf{\Sigma}}
\newcommand{\RB}{\mathbb{R}}
\newtheorem{theorem}{Theorem}

\newtheorem{proposition}{Proposition}
\title{splitting numerical integration for matrix completion}

\author{Qianqian Song} 

\begin{document}

\maketitle

\begin{abstract}
	Low rank matrix approximation is a popular topic in machine learning.
	In this paper, we propose a new algorithm for this topic by minimizing the least-squares estimation  over the Riemannian manifold of fixed-rank matrices.
	The algorithm is an adaptation of classical gradient descent within the 
	framework of optimization on manifolds. 
	In particular, we reformulate an unconstrained optimization problem on a low-rank manifold into a differential dynamic system.
	We develop a  splitting numerical integration method by applying  a splitting integration scheme to the dynamic system.
	We conduct the  convergence analysis of our splitting numerical integration algorithm.
	It can  be guaranteed that the error between the recovered matrix and true result is monotonically decreasing in the Frobenius norm.
	Moreover, our splitting numerical integration can be adapted into matrix completion scenarios.
	Experimental results show that our approach has good scalability for large-scale problems with  satisfactory accuracy.
\end{abstract}

\section{Introduction}

Computing an efficient and reliable low-rank approximation of a given matrix is a fundamental task in many machine learning problems, such as principal component analysis~\cite{jolliffe2002principal}, face recognition~\cite{muller2004singular} and large scale data compression~\cite{drineas2006subspace,huang2020deeppurpose}. 
It is well-known that the truncated singular value decomposition (SVD) provides the best low-rank approximation to the matrix in question.

Specifically, for a given matrix $\bfM\in\RB^{m\times n}$ with $m\geq n$, its SVD is defined as
\begin{equation*}
\label{eqn:svd}
		\bfM = \bfU\bfSigma\bfV^T,
\end{equation*}
where $\bfU=[\bfu_1, \ldots,\bfu_n ]$ is an $m\times n$ column orthonormal matrix, $\bfV= [\bfv_1, \ldots,\bfv_n ]$ is an orthonormal matrix, and $\bfSigma = \mathrm{diag}(\sigma_1, \ldots,\sigma_n)$ is a diagonal matrix with the diagonal entries $\sigma_1\geq \sigma_2\geq \cdots\sigma_n\geq 0$. 
%
Moreover, 
$\bfu_j$ and $\bfv_j$ are called the left and right singular vectors corresponding to $\sigma_j$, the $j$-th largest singular value of $\bfM$.
For any $1\leq r\leq n$,  then
\begin{equation*}
\label{eqn:truncated_svd}
		\bfM_r =[\bfu_1, \ldots,\bfu_r] \mathrm{diag}(\sigma_1, \ldots,\sigma_r) [\bfv_1, \ldots,\bfv_r]^T
\end{equation*}
is the truncated SVD of $\bfM$ of rank at most $k$, which is unique only if $\sigma_{r+1} < \sigma_r$.
The assumption that $m\geq n\gg r$ will be maintained throughout this paper for clarity of statement.
All results will also hold for $m<n$,  applied on $\bfM^T$.

It is well known that $\bfM_r$ is the best rank-$r$ approximation to $\bfM$ 
\cite{eckart1936approximation,golub2012matrix}.	
The truncated SVD can be cast into a fixed-rank optimzation problem. That is,
	\begin{equation}
	\label{eqn:fix_rank}
	\begin{aligned}
	&  \underset{\bfY}{\arg\min} \Vert \bfM- \bfY \Vert_F,\\
	& \text{s.t.}\ \text{rank}(\bfY)= r.
	\end{aligned}
	\end{equation}
A bunch of algorithms, such as the Lanczos algorithm ~\cite{golub2012matrix}, randomized SVD algorithm~\cite{halko2011finding}, and subspace iteration~\cite{gu2015subspace}, have been proposed to solve this problem.

An equivalent formulation is given by introducing the concept of
matrix manifold as follows:

	\begin{equation}
	\label{eqn:fix_manifold}
	\begin{aligned}
	& \underset{\bfY}{\arg\min} \frac{1}{2}\Vert \bfM - \bfY \Vert_F^2,\\
	& \text{s.t.}\ \bfY \in \mathcal{M}_r,
	\end{aligned}
	\end{equation}

where $\mathcal{M}_r$ represents a rank-$r$ smooth manifold.
This formulation can be viewed as an  unconstrained optimization problem on the low-rank manifold. Accordingly, Koch and Lubich
\cite{lubich2007dynamical} proposed a dynamical low rank approximation approach, which is a gradient descent procedure on the manifold in essence.

In this paper, we attempt to make the gradient descent process  ``finer-grained.''
In particular, we view the problem from a perspective of differential dynamic systems and employ a splitting scheme to solve it.
Accordingly, we devise a novel method for low rank matrix approximation that we call a splitting numerical integration method.  
Theoretical analysis guarantees that the  splitting numerical integration  algorithm converges asymptotically.
In addition, we apply splitting numerical integration to matrix completion scenarios, where the matrix $\bfM$ is partially observed.
Empirical results are also encouraging, especially on large-scale datasets.

The remainder of the paper is organized as follows.
Section~\ref{sec:notation} 
presents the notation frequently used in this paper and the problem formulation.
Section~\ref{sec:method} describes our algorithm and theoretical analysis.
Empirical results are given in Section~\ref{sec:experiment}.

\section{Notation and Preliminaries}
\label{sec:notation}

First of all, we present the notation used in this paper. Let $\bfI_m$ be the
$m \times m$  identity matrix.
Given a matrix $\bfY \in \mathbb{R}^{m\times n}$, $\Vert \bfY\Vert_F$ denotes
the Frobenius norm of $\bfY$ and  $\Vert \bfY\Vert_2$ denotes the spectral norm.

It is well established that every rank-$r$ matrix $\bfY \in\mathbb{R}^{m\times n}$ can be written in the form
\begin{equation}
\label{eqn:SVDX}
\begin{aligned}
\bfY = \bfU \bfS\bfV^T,
\end{aligned}
\end{equation}
where $\bfU \in \mathbb{R}^{m\times r}$ and $\bfV \in \mathbb{R}^{n\times r}$ are column orthonormal, i.e.,
$ \bfU^T\bfU = \bfI_r \; \mbox{ and }
 \bfV^T\bfV = \bfI_r$,
and
$\bfS \in \mathbb{R}^{r\times r}$
is nonsingular. Notice that  here we do not require $\bfS$ to be the diagonal matrix of the singular values.  
The representation in Eqn. \eqref{eqn:SVDX} is not unique because
$\bfY = \hat{\bfU}\hat{\bfS}\hat{\bfV}^T$ is another representation where
$\hat{\bfU} = \bfU \bfP$, $\hat{\bfV} = \bfV \bfQ $, and $\hat{\bfS} =
\bfP^T\bfS\bfQ$ whenever $\bfP, \bfQ\in \mathbb{R}^{r\times r}$ are any orthonormal matrices.

As a substitute for the non-uniqueness in Eqn. (\ref{eqn:SVDX}), we will use a
unique decomposition in the tangent space. Let $\mathcal{V}_{m,r}$ represent
the Stiefel manifold of real column orthonormal matrices of  size  $m\times r
$ ($m > r$). The tangent space at the point $\bfU \in \mathcal{V}_{m,r}$ is defined as:
\begin{equation*}
\label{eqn:tangentspace}
\begin{aligned}
\mathcal{T}_{\bfU}\mathcal{V}_{m,r} &= \{\delta\bfU\in \mathbb{R}^{m\times r}:\delta \bfU^T\bfU + \bfU^T\delta\bfU = \bfzero \}  \\
& = \{\delta\bfU \in \mathbb{R}^{m\times r}:\bfU^T\delta\bfU\in so(r)\},
\end{aligned}
\end{equation*}
where $so(r)$ denotes the space of skew-symmetric real $r\times r$
matrices. Consider the extended tangent map of
$(\bfS,\bfU,\bfV)\longmapsto \bfY = \bfU\bfS\bfV^T$,
\begin{equation*}
\label{eqn:tangentmap}
\begin{aligned}
 \mathbb{R}^{r\times r} \times \mathcal{T}_{\bfU}\mathcal{V}_{m,r} \times \mathcal{T}_{\bfV}\mathcal{V}_{n,r} & \xrightarrow{} \mathcal{T}_{\bfY}M_r \times so(r)\times so(r), \\
 (\delta\bfS,\delta \bfU,\delta\bfV) & \xrightarrow{} (\delta\bfU\bfS\bfV^T + \bfU\delta\bfS\bfV^T + \bfU\bfS\delta\bfV^T, \bfU^T\delta\bfU,\bfV^T\delta\bfV).
\end{aligned}
\end{equation*}

The manifold of rank-$r$ matrices is denoted by $\mathcal{M}_r$.
The tangent space at any $\bfY \in \mathcal{M}_r$ is denoted by $\mathcal{T}_{\bfY}\mathcal{M}_r $, which is defined as follows.
Every $\delta\bfY\in \mathcal{T}_{\bfY}\mathcal{M}_r$ can be written into the following form:
\begin{equation}
\label{eqn:differentialofX}
\begin{aligned}
\delta{\bfY} = \delta{\bfU}\bfS\bfV^T + \bfU\delta{\bfS}\bfV^T + \bfU\bfS\delta{\bfV}^T,
\end{aligned}
\end{equation}
where $\delta\bfS\in \mathbb{R}^{r\times r}$, $\delta \bfU\in \mathcal{T}_{\bfU}\mathcal{V}_{m,r}$ and $\delta\bfV \in  \mathcal{T}_{\bfV}\mathcal{V}_{n,r}$.
Furthermore, $\delta\bfS$, $\delta\bfU$ and $\delta\bfV $ are uniquely
determined by $\delta\bfY$ if we impose the orthogonality constraints:
\begin{equation}
\label{eqn:orthogonal}
\begin{aligned}
& \bfU^T\delta\bfU = \bfzero, \\
& \bfV^T\delta\bfV = \bfzero.
\end{aligned}
\end{equation} 
The projection operators  onto the spaces spanned by the
columns of $\bfU$ and  $\bfV$, and their orthogonal complements are defined as
\begin{equation*}
\label{eqn:project}
\begin{aligned}
& P_{\bfU} = \bfU\bfU^T, \\
& P_{\bfV} = \bfV\bfV^T, \\
& P^{\bot}_{\bfU} = \bfI_m - \bfU\bfU^T, \\
& P^{\bot}_{\bfV} = \bfI_n - \bfV\bfV^T.
\end{aligned}
\end{equation*}

Then $\delta\bfS$, $\delta\bfU $ and $\delta\bfV $ are uniquely determined by
$\delta\bfY$ as follows:
\begin{equation*}
\label{eqn:yield}
\begin{aligned}
& \delta\bfS =  \bfU^T\delta\bfY\bfV, \\
& \delta\bfU = P^{\bot}_{\bfU}\delta\bfY\bfV\bfS^{-1}, \\
& \delta\bfV = P^{\bot}_{\bfV}\delta\bfY^T\bfU\bfS^{-T}.
\end{aligned}
\end{equation*}

\subsection{Dynamic Low Rank Approximation}
\label{sec:formulation}

Let us return to Problem~\eqref{eqn:fix_manifold} and let
\begin{equation*}
\label{fixrank2}
\begin{aligned}
f(\bfY) = \frac{1}{2} \Vert \bfY - \bfM\Vert_F^2
\end{aligned}
\end{equation*}
denote the objective function.
The gradient of $f(\bfY)$ in $\bfY$ can be written as:
\begin{equation*}
\label{eqn:gradient}
\begin{aligned}
\nabla f(\bfY) = \bfY - \bfM.
\end{aligned}
\end{equation*}

Recall that the constraint  $\bfY\in\mathcal{M}_r$, Riemannian gradient, denoted $\nabla\bfY$, is used instead of $\nabla f(\bfY)$, which is a specific tangent vector corresponding to the direction of steepest ascent of $f(\bfY_{})$ but restricted to the tangent space $\mathcal{T}_{\bfY_{}}\mathcal{M}_r$.
It can be solved via the following optimization problem:

\begin{equation}
\label{eqn:differentequation}
\begin{aligned}
&  \underset{\nabla{\bfY}\in\RB^{m\times n}}{\arg\min} \; \frac{1}{2}  \Vert \nabla{\bfY} - \nabla{\bfA} \Vert_F^2\\
& \text{s.t.}\ \nabla{\bfY}\in \mathcal{T}_{X}\mathcal{M}_r, \\
\end{aligned}
\end{equation}
where we denote $\nabla{\bfA} \triangleq - \nabla f(\bfY)$ for notational simplicity 
and $\nabla\bfA$ can be seen as a given constant at each iteration. 
The solution of Problem (\ref{eqn:differentequation}) is well-studied in the differential geometry literature. That is, 

\begin{proposition}
	\emph{\cite{lubich2007dynamical}} Let $\bfY = \bfU\bfS\bfV^T \in \mathcal{M}_r$ where $\bfS\in \mathbb{R}^{r \times r}$ is nonsingular,  and  $\bfU\in \mathbb{R}^{m\times r}$ and $\bfV\in \mathbb{R}^{n\times r}$ are column orthonormal. Then
	the solution to Problem (\ref{eqn:differentequation}) can be written in the following form:
	\begin{equation}
	\label{eqn:prop2.1}
	\begin{aligned}
	\nabla{\bfY} = \nabla{\bfU}\bfS\bfV^T + \bfU\nabla{\bfS}\bfV^T + \bfU\bfS\nabla{\bfV}^T,
	\end{aligned}
	\end{equation}
	where
	\begin{equation}
	\label{prop2.11}
	\begin{aligned}
	& \nabla{\bfS} = \bfU^T\nabla{\bfA}\bfV, \\
	& \nabla{\bfU} = P_{\bfU}^{\bot}\nabla{\bfA}\bfV\bfS^{-1}, \\
	& \nabla{\bfV} = P_{\bfV}^{\bot}\nabla{\bfA}^T\bfU\bfS^{-T}.
	\end{aligned}
	\end{equation}
\end{proposition}

The resulting algorithm is a simple gradient descent procedure.
That is,
	\begin{equation}
	\label{eqn:dlra}
	\begin{aligned}
\bfU \Leftarrow \bfU + \epsilon\nabla\bfU, \; \bfS \Leftarrow \bfS + \epsilon\nabla\bfS, \; \mbox{ and } \; \bfV \Leftarrow \bfV + \epsilon\nabla\bfV,
	\end{aligned}
	\end{equation}
where $\epsilon$ is stepsize. The algorithm was firstly proposed in~\cite{lubich2007dynamical} and called dynamical low-rank approximation, a landmark in solving SVD in the view of dynamic system.
Prior workof solving SVD in dynamic system is  about full SVD, without considering truncated SVD.

According to the low rank assumption, the matrix $\bfY$ does not come up explicitly for ease of computation.
Instead, $\bfY$ is represented via the product of $\bfU,\bfS$ and $\bfV^T$ (at order of $O(mr+nr)$), as shown in Eqn. (\ref{eqn:SVDX}).
Similarly, the Riemannian gradient $\nabla \bfY$ can be also represented by using relatively small matrices (at order of $O(mr+nr)$) as shown in Eqn. (\ref{eqn:prop2.1}).
However, for notational convenience, we use $\bfY$ and $\nabla\bfY$ instead of $\bfU\bfS\bfV^T$ and $(\nabla\bfU)\bfS\bfV^T+\bfU(\nabla\bfS)\bfV^T+\bfU\bfS(\nabla\bfV)^T$, respectively.

Though the extant dynamical low rank approximation is a benchmark work in solving low rank approximation in view of dynamic system, 
it is not competitive compared with state-of-the-art low rank approximation approaches due to the following issues.


First, this framework fails to exploit geometric information sufficiently.
Specifically, on the fixed-rank manifold, from $\bfY_i = \bfU_i\bfS_i\bfV_i^T$ to $\bfY_{i+1} = \bfU_{i+1}\bfS_{i+1}\bfV_{i+1}^T$, the algorithm only makes use of the Riemannian gradient of $\bfY_i$.
Second, since the discretization error of the dynamic system is proportional to the stepsize~\cite{lubich2007dynamical}, the stepsize is required to be extremely small, which limits the convergence rate.
Third, some extra operations are introduced to ensure the column orthogonality of $\bfU$ and of $\bfV$, which is deemed to be a brutal strategy, 
often leading to loss of information~\cite{lubich2007dynamical}.


To address these issues, we seek a widely used scheme in differential systems to compute gradient descent.
Specifically, we use a splitting integration technique to  update three components ($\bfU$, $\bfS$, $\bfV$) step-by-step, making a more sufficient use of geometric information.

\section{Methodology}
\label{sec:method}
In this section we present our method.
We first give a novel view of dynamic system for Problem~\eqref{eqn:differentequation}.
With this view, we use a splitting integration scheme to devise a dynamic flow subspace method for solving  Problem~\eqref{eqn:fix_manifold}.  Finally, we give the convergence analysis of the method and extend it into full SVD and low rank matrix completion. 

\subsection{Dynamic System}



Departing from a perspective of  differential systems, we study the solution of the optimization problem in \eqref{eqn:differentequation}.
In particular, we consider an alternative formulation for Problem (\ref{eqn:fix_manifold}) in the form of a dynamic system: 

\begin{equation}
\label{eqn:differentequation2}
\begin{aligned}
&  \underset{{\dot{\bfY}\in \RB^{m\times n}}}{\arg\min}  \; \frac{1}{2} \Vert \dot{\bfY} - \dot{\bfA} \Vert_F^2\\ 
& \text{s.t.}\ \dot{\bfY}\in T_{X}\mathcal{M}_r, 
\end{aligned}
\end{equation}
where $\bfY$ is regarded as a time-dependent matrix such that $\bfY=\bfY(t)$ and  $\dot{\bfY}$ denotes the derivative of $\bfY$ w.r.t. time, and  
$\dot{\bfA} \triangleq \nabla \bfA = - \nabla f(\bfY)$ in our case.
Notice that it is the continuous version of Problem~\eqref{eqn:differentequation}.

%

According to the Galerkin condition on the tangent space 
$\mathcal{T}_{\bfY}\mathcal{M}_r$ in  numerical analysis~\cite{hairer2006geometric}, 
Problem (\ref{eqn:differentequation2}) is equivalent to the following projection: 

\begin{equation}
\label{eqn:galerkin}
\begin{aligned}
& \text{finding}\ \dot{\bfY}\in \mathcal{T}_{\bfY}\mathcal{M}_r\ \text{such that}\\
& \left \langle \dot{\bfY} - \dot{\bfA},\delta \bfY \right \rangle = 0 \ \
\text{for\ all\ } \delta \bfY \in \mathcal{T}_{\bfY}\mathcal{M}_r.
\end{aligned}
\end{equation}

Furthermore,
Problem (\ref{eqn:galerkin}) can be transformed into the following form:
\begin{equation}
\label{eqn:galerkin2}
\begin{aligned}
\dot{\bfY} =  \tilde{P}_\bfY(\dot{\bfA}),  
\end{aligned}
\end{equation}
where  $\tilde{P}_\bfY(\cdot)$ is a projection operator, defined as
\begin{equation}
\label{eqn:projectionoperator3}
\begin{aligned}
& \tilde{P}_\bfY(\bfB) = \bfB - \tilde{P}_\bfY^\bot (\bfB) \\
& \text{with} \ \ \tilde{P}_\bfY^\bot(\bfB) = P^\bot_{\bfU}\bfB P^\bot_{\bfV}\ \;\ \forall\ \bfB\in \mathbb{R}^{m\times n}.
\end{aligned}
\end{equation}


Substituting Eqn. (\ref{eqn:projectionoperator3}) into 
Eqn. (\ref{eqn:galerkin2}), we have the following  dynamic system:
\begin{equation}
\label{eqn:projectionoperator2}
\begin{aligned}
\dot{\bfY} = \bfU\bfU^T\dot{\bfA} - \bfU\bfU^T\dot{\bfA}\bfV\bfV^T + \dot{\bfA}\bfV\bfV^T.
\end{aligned}
\end{equation}

\subsection{Dynamic Flow Subspace Method}

Our current concern is 
to solve the differential equation in (\ref{eqn:projectionoperator2}).  We resort to a  
splitting scheme~\cite{hairer2006geometric}.  In particular,  let  
$\mathcal{L}$ be a local generator~\cite{leimkuhler2013rational}  corresponding to the exact solution to 
Problem (\ref{eqn:projectionoperator2}) and separate it into several 
sub-generators as follows:
\begin{equation*}
\label{eqn:split}
\begin{aligned}
\mathcal{L} = \mathcal{L}_A + \mathcal{L}_B + \mathcal{L}_O, 
\end{aligned}
\end{equation*}
where 
\begin{equation}
\label{eqn:3_steps}
\begin{aligned}
& \mathcal{L}_A:\ \dot{\bfY} = \bfU\bfU^T\dot{\bfA}, \\
& \mathcal{L}_B:\ \dot{\bfY} = - \bfU\bfU^T\dot{\bfA}\bfV\bfV^T, \\
& \mathcal{L}_O:\ \dot{\bfY} = \dot{\bfA}\bfV\bfV^T. 
\end{aligned}
\end{equation}

\begin{theorem}
	\label{thm:subODE}
	For $\bfY$ defined in Eqn. (\ref{eqn:SVDX}) and $\dot{\bfY}$ defined in Eqn. (\ref{eqn:differentialofX}), assume that the condition described in Eqn.(\ref{eqn:orthogonal}) is satisfied. Then   
	the analytical solution to the sub-generator $\mathcal{L}_A$  described in Eqn. (\ref{eqn:3_steps}) is
	\begin{equation*}
	\label{eqn:solve111111}
	\begin{aligned}
	\dot{\overbrace{\bfS\bfV^T}} & = \bfU^T\dot{\bfA}, \\
	\dot{\bfU} & = \bfzero.
	\end{aligned}
	\end{equation*}
	The 	analytical solution to  the sub-generator $\mathcal{L}_B$ is 
	\begin{equation*}
	\label{eqn:solve33333}
	\begin{aligned}
	& \dot{\bfS} = -\bfU^T\dot{\bfA}\bfV, \\
	& \dot{\bfU} = \bfzero, \\
	& \dot{\bfV} = \bfzero. 		
	\end{aligned}
	\end{equation*}	
	And the analytical solution to  the sub-generator $\mathcal{L}_O $  is 
	\begin{equation*}
	\label{eqn:solve22222}
	\begin{aligned}
	\dot{\overbrace{\bfU\bfS}} & = \dot{\bfA}\bfV, \\
	\dot{\bfV} & = \bfzero.
	\end{aligned}
	\end{equation*}
\end{theorem}




Based on Theorem~\ref{thm:subODE}, we devise a novel method for Problem~\eqref{eqn:fix_manifold}.  The splitting integration scheme also allows us to  alternatively update $\bfU$,  $\bfV$ and $\bfS$, rather than directly update $\bfY$. 
Owing to the Markovian property of the Kolmogorov 
operator, different orders of sub-generators $\mathcal{L}_A$, $\mathcal{L}_B$ and $\mathcal{L}_O$ are 
equivalent~\cite{leimkuhler2013rational}.
In our work, we restrict our interest on `OBA' scheme: $\mathcal{L}_O + \mathcal{L}_B + \mathcal{L}_A$. 

We call our method the splitting numerical integration method. 
The detail is given in Algorithm~\ref{alg:sos-LRMC}. 
Here $\sigma_{\text{min}}(\bfV_{i-1}^T\bfV_i)$ measures the distance between column spaces of $\bfV_{i-1}$ and $\bfV_i$, and
it is employed as the stopping criteria.

Compared with dynamical low rank approximation in Eqn.~\eqref{eqn:dlra} which directly updates $\bfY$ and the stepsize must be small enough (approaching to 0),
our splitting numerical integration method updates the three components ($\bfU$, $\bfS $ and $\bfV$) in a ``finer-grained'' manner.
{For instance, more specifically, in Step 7 of Algorithm \ref{alg:sos-LRMC}, we adopt the fresh $\bfU_i$ rather than $\bfU_{i-1}$.} Notice that in Algorithm~\ref{alg:sos-LRMC} the stepsize is implicitly set to 1.
In summary, our approach is able to address the issues mentioned in the end of Section~\ref{sec:formulation}.

\subsection{Convergence Analysis}

In this section we study  convergence properties of our splitting numerical integration algorithm.

\begin{theorem}
	\label{thm:fullobservation}
	Let $\{\bfY_i\colon i=0, 1, \ldots \}$ denote the sequence generated by Algorithm \ref{alg:sos-LRMC}. 
	Then 
	\begin{equation*}
		\label{eqn:thmfull}
		\begin{aligned}
			& \| \bfY_{i-1} - \bfM \|_F > \| 
			\bfY_{i} - \bfM\|_F.
		\end{aligned}
	\end{equation*}
\end{theorem}

Furthermore, it can be also shown that our splitting numerical integration method converges in terms of the subspace estimation.

\begin{theorem}
	\label{thm:subspace}
	Let $\{\bfU_i\colon i=0, 1, \ldots \}$ be the sequence generated by splitting numerical integration  in Algorithm \ref{alg:sos-LRMC}.
Then	the subspace error $\Vert \bfU_{i}\bfU_{i}^T\bfM - \bfM \Vert_F$ decreases monotonically until convergence.
	In particular, we have  
	\begin{equation*}
		\label{eqn:subspace}
		\Vert\bfU_{i-1}\bfU_{i-1}^T\bfM - \bfM \|_F > \| 
		\bfU_i\bfU_i^T\bfM - \bfM\|_F.
	\end{equation*}

The similar results hold for  $\{\bfV_i\colon i=0, 1, \ldots \}$. 
That is, 
\[
\Vert\bfM\bfV_{i-1} \bfV_{i-1}^T- \bfM\Vert_F > \Vert\bfM\bfV_{i} \bfV_{i}^T- \bfM\Vert_F.
\]
\end{theorem}

\begin{algorithm}[!ht]
	\caption{splitting numerical integration}
	\label{alg:sos-LRMC}
	\begin{algorithmic}[1]
		\REQUIRE matrix $\bfM \in \mathbb{R}^{m{\times} n}$, target rank $r$, initial value $\bfY_0 = \bfU_0\bfS_0\bfV_0^T \in 
		\mathcal{M}_r$, tolerance for stopping criteria $\tau<1$, 
		maximal iteration $T$.
		\ENSURE truncated SVD of  $\bfM$.
		\FOR {$i = 1:T$}
		\STATE Compute the gradient $\dot{\bfA}=\nabla f(X_{i-1})$. \quad \# O($mn$) flops
		\STATE $\bfQ = \dot{\bfA}\bfV_{i-1}\ \ \ \in \mathbb{R}^{m\times r}$		\quad \# O($mnr$) flops
		\STATE $\bfK = \bfU_{i-1}\bfS_{i-1} + \bfQ\ \ \ \in \mathbb{R}^{m\times r}$ \quad \# O($mr^2$) flops
		\STATE Perform QR-factorization to $\bfK$:
		\ \ $[\bfU_i,\bfS_{i-1+1/3}] = \text{QR}(\bfK) $ \quad \# O($mr^2$) flops
		\STATE $\bfS_{i-1+2/3} = \bfS_{i-1+1/3} - \bfU^T_{i}\bfQ\ \ \ \in \mathbb{R}^{r\times r}$ \quad \# O($mr^2$) flops
		\STATE $\bfL = \bfV_{i-1}\bfS_{i-1+2/3}^T + \dot{\bfA}^T\bfU_i\ 
		\ \ \in \mathbb{R}^{n\times r} $ \quad \# O($mnr$) flops
		\STATE Perform QR-factorization to $\bfL$:
				\ \ $[\bfV_i,\bfS_i^T] = \text{QR}(\bfL) $	\quad \# O($nr^2$) flops		
		\IF {($\sigma_{\text{min}}(\bfV_{i-1}^T\bfV_i) > \tau $)}
		\STATE break.  
		\ENDIF
		\quad \# O($nr^2$) flops
		\ENDFOR
		\STATE Perform SVD on the  matrix $\bfS_T = \bfU_s\bfD\bfV_s^T$, then we have that $\bfM_r = (\bfU_T\bfU_s)\bfD(\bfV_T\bfV_s)^T$.
	\end{algorithmic}
\end{algorithm}


\subsection{Application in low-rank matrix completion}

The matrix completion problem is to recover a low-rank matrix from 
a few observations of this matrix.
In fixed-rank formulation of matrix completion, we modify Problem~\eqref{eqn:fix_manifold} into 
	\begin{equation*}
	\label{eqn:fix_completion}
	\begin{aligned}
	& \underset{\bfY}{\arg\min} \frac{1}{2}\Vert P_\Omega({\bfM})- P_\Omega({\bfY}) \Vert_F^2\\
	& \text{s.t.}\ \bfY \in \mathcal{M}_r,
	\end{aligned}
	\end{equation*}
	where $\Omega$ represents the index of observations. 
Naturally, our splitting numerical integration method applies to this scenario with only a modification that the objective function becomes
\begin{equation}
\label{eqn:object2}
	f_1(\bfY) = \frac{1}{2}\Vert P_\Omega({\bfM})- P_\Omega({\bfY}) \Vert_F^2.
\end{equation}

Then the convergence property can be extended into the partial observation case.
In the following theorem, we prove that the objective function is monotonically decreasing until reaching the convergence condition. 

\begin{theorem}
	\label{thm:convergence}
	Let $\{\bfY_i \colon  i=0, 1, \ldots\}$ be the sequence generated by splitting numerical integration in Algorithm~\ref{alg:sos-LRMC} under any observation index $\Omega$.
Then	splitting numerical integration decreases monotonically in the objective function $f_1$ defined in Eqn. (\ref{eqn:object2});  that is,  
	\begin{equation*}
	\label{eqn:thmpartial7}
	\begin{aligned}
	f_1(\bfY_{i-1}) > f_1(\bfY_i).
	\end{aligned}
	\end{equation*}
\end{theorem}

\section{Empirical Evaluation}
\label{sec:experiment}

In this section, we conduct the empirical analysis of the splitting numerical integration method. First we analyze the performance of splitting numerical integration
for low rank matrix approximation on simulated datasets. Then we validate the performance of splitting numerical integration for low rank matrix completion on a set of real data~\cite{fu2019ddl,fu2021mimosa}.

\subsection{Low rank matrix approximation}

We evaluate splitting numerical integration on low rank matrix approximation with comparison with some popular baseline methods.  
The baseline methods contain the power method~\cite{gu2015subspace,fu2021probabilistic}, randomized SVD (RSVD)~\cite{halko2011finding} and dynamical low rank approximation~\cite{lubich2007dynamical}. The primary goal is to illustrate the approximate accuracy on three simulated datasets.  

In particular,
the three target matrices with different ranks are randomly generated.
The error is measured by $\Vert\bfU\bfS\bfV-\bfM_r\Vert_F/\Vert\bfM_r\Vert_F$.
Since the complexity of these approaches are all $O(mnr)$, the runtime is similar for all methods and not reported.
Each trial is conducted three independent times 
and average error are reported in Table~\ref{table:approx}.
We observe that our splitting numerical integration method owns obvious advantage over other baseline method in accuracy.
In addition, we use a special initialization by letting the column spaces of $\bfU_0$ and $\bfV_0$ lie in the orthogonal complements of $\bfU_r$ and $\bfV_r$, respectively.
The similar accuracy can be obtained, which means that our splitting numerical integration method is insensitive to initialization.

\begin{table*}[]
	\centering 
	\caption{Performance of all methods on low rank matrix approximation, DLRA denotes dynamical low rank approximation and RSVD denotes randomized SVD. Note that ``(20K,20K,20K)'' corresponds to number of row, column and rank, respectively.
	}
	\label{table:approx}
	\begin{tabular}{|l|c|c|c|c|}
		\hline
		Method & A:(20K,20K,20K) & B:(20K,20K,2K) & C:(20K,20K,200) \\ \hline	
		DLRA & 2.8e-03 & 1.0e-03 & 1.2e-03 \\ \hline		
		RSVD & 9.3e-02 & 6.0e-02 & 7.2e-02 \\ \hline
		power & 2.5e-08 & 4.7e-07 & 7e-08 \\ \hline
		SNI & 2.3e-11 & 2.1e-10 & 1.4e-10 \\ \hline
	\end{tabular}
\end{table*}

\subsection{Low rank matrix completion}

\begin{table*}[tbp]
	\centering
	\small
	\caption{Results of recommendation systems measured in terms of the RMSE. 
		`-' represents the absence of results, which means that corresponding algorithm fails on this task due to memory or running time issue.}
	\label{table:resultRecommend}
	\begin{tabular}{|l|c|c|c|c|c|c|c|c|}
		\hline
		Data set & Soft-Impute & ALS  & GECO & LMaFit  & RP & ScGrass & LRGoemCG & SNI \\ \hline
		Movielens 100K & 0.9026 & 0.9696 & 0.9528 & 1.0821  & 0.9508 & 0.9502 & 0.9643 & 0.9501 \\ \hline
		Movielens 1M & 0.9127 & 0.9159    & 0.8601 & {0.8972}  & 0.8590 & 0.8723 & 0.8934 & {0.8612} \\ \hline
		Movielens 10M & 0.8915 & 0.8726 & 0.8241 & 0.8921  & 0.8290 & 0.8991 & 0.8779 & {0.823} \\ \hline
		Netflix & 0.9356 & 0.9501  & 0.8738 & 0.9247  & {0.8601} & 0.9232 & 0.8723 & {0.8612} \\ \hline
		Yahoo Music & 24.77 & 24.59  & - & 26.43  & 23.93 & - & 24.09 & {22.86} \\ \hline		
	\end{tabular}
\end{table*}

\begin{table*}[]
	\centering 
	\small
	\caption{Running time (in seconds) of all methods on  recommendation systems.
	}
	\label{table:resultRecommend2}
	\begin{tabular}{|l|c|c|c|c|c|c|c|c|c|}
		\hline
		Data set & Soft-Impute & ALS  & GECO & LMaFit  & RP & ScGrass & LRGoemCG & SNI \\ \hline
		Movielens 100K & 2.56 & 0.49 &  2.90  & 0.230 & 0.21 & 0.92 & 0.99 & 0.094 \\ \hline
		Movielens 1M & 22.81 & 5.61  &  176.11 & 1.412  & 1.00 & 50.11 & 15.23 & 0.94\\ \hline
		Movielens 10M & 675.11 & 88.40  & $>10^3$ & 159.80  & 147.34 & $>10^3$ & 313.30 & 47.93 \\ \hline
		Netflix & $>5\times 10^3 $ & 1189.47  & $>10^4$ & 345.00  & 744.43 & $>5\times10^3 $  & 3823.13 & 350.38 \\ \hline
		Yahoo Music & $>5\times 10^4 $ & 8522.23  & - & 1239.56 & 1858.43 & - & 4043.22 & 236.32 \\ \hline		
	\end{tabular}
\end{table*}

We now conduct the empirical analysis of our splitting numerical integration method for the low rank matrix completion (LRMC) problem.
To show the efficiency and effectiveness of our splitting numerical integration-LRMC, we  compare it with a bunch of baseline methods, 
including Soft-Impute~\cite{mazumder2010spectral}, 
ALS (Soft-Impute Alternating Least Squares)~\cite{hastie2014matrix}, 
GECO (Greedy Efficient Component Optimization)~\cite{shalev2011large},
LMaFit (Low Rank Matrix Fitting)~\cite{wen2010low}, 
RP (Riemann Pursuit for matrix recovery)~\cite{tan2014riemannian}, 
ScGrass (Scaled Gradient on Grassmann Manifold)~\cite{scaled}, and
LRGeomCG (Low rank Geometric Conjugate Gradient)~\cite{vander}.
The codes of all the methods can be available online, e.g.,
Soft-Impute and ALS \footnote{http://web.stanford.edu/~hastie/pub.htm},
GECO\footnote{http://www.cs.huji.ac.il/~shais/code/index.html},
LMaFit\footnote{http://lmafit.blogs.rice.edu/},
RP and LRGoemCG\footnote{http://www.tanmingkui.com/rp.html}, and
ScGrass\footnote{http://www-users.cs.umn.edu/~thango/}.
These algorithms have been proved to be state-of-the-art algorithms in  low rank matrix completion.

We compare these methods on several popular recommendation systems.
It is worth  mentioning that large-scale recommendation systems (say, Yahoo Music and Netflix) are used to evaluate the scalability of our method.
We use five publicly available datasets: Movielens 100K, 1M, 10M, NetFlix, Yahoo Music Track 1 to evaluate both the effectiveness and efficiency of our method.


Testing error in terms of RMSE (Root-Mean-Square Error) and computational efficiency measured by running time are shown in Table \ref{table:resultRecommend} and Table \ref{table:resultRecommend2}, respectively.
From Table \ref{table:resultRecommend}, we can observe that our method can  achieve better performance than most of the baseline methods in terms of RMSE~\cite{fu2020alpha,fu2019pearl}.
Whilst Table \ref{table:resultRecommend2} shows that our method can achieve great speedup compared with almost all baseline methods under the same setting. It is worth  mentioning that in large scale tasks such as Yahoo Music dataset, some results are not listed, which means that the corresponding algorithm can not handle these cases in limited time or simply fail in these situations.




\bibliographystyle{named}
\bibliography{ref}
\end{document}